# Leveraging CNNs and Ensemble Learning for Automated Disaster Image Classification

Archit Rathod[1*], Veer Pariawala[1*], Mokshit Surana[1*], and Dr. Kumkum Saxena[1]

[1] Thadomal Shahani Engineering College, Mumbai
{architrathod77, pariawalaveer, mokshitsurana3110}@gmail.com
kumkum.saxena@thadomal.org

[*] These authors contributed equally to this work.

**Abstract.** Natural disasters act as a serious threat globally, requiring effective and efficient disaster management and recovery. This paper focuses on classifying natural disaster images using Convolutional Neural Networks (CNNs). Multiple CNN architectures were built and trained on a dataset containing images of earthquakes, floods, wildfires, and volcanoes. A stacked CNN ensemble approach proved to be the most effective, achieving 95% accuracy and an F1 score going up to 0.96 for individual classes. Tuning hyperparameters of individual models for optimization was critical to maximize the models' performance. The stacking of CNNs with XGBoost acting as the meta-model utilizes the strengths of the CNN and ResNet models to improve the overall accuracy of the classification. Results obtained from the models illustrated the potency of CNN-based models for automated disaster image classification. This lays the foundation for expanding these techniques to build robust systems for disaster response, damage assessment, and recovery management.

**Keywords:** CNNs, ResNet, BERT, XGBoost, ReLU, Ensemble Modeling

## 1 Introduction

Disasters are a major part of human life having devastating effects worldwide. Classification of natural disasters is critical to disaster management and its response. The different types and characteristics of natural disasters are pivotal to minimizing their impact on human life and saving lives.

An important part of disaster management is its identification or classification since it provides a framework for responding to the effects of the different types of natural disasters. Natural disasters such as hurricanes, floods, wildfires, and earthquakes cause severe damage and destruction around the world affecting millions of people each year.

In recent years, there have been various machine learning techniques such as convolutional neural networks (CNNs) [1], that have become powerful tools for automated hazard classification. The study and experiments evaluate the use of CNNs in natural disaster classification and provide detailed information on the various types of natural disasters and their characteristics.

Natural disasters can strike unexpectedly, endangering people and causing significant damage to human life. Classification of these disasters is important since it helps government authorities to plan, respond, and relieve the effects of natural disasters. Understanding the characteristics of these different natural hazards would help strategies to be developed to control and mitigate their effects.

## 2 Literature Review

Natural disasters are frequently classified based on factors such as their origin, intensity, and impact on human lives and ecosystems. Techniques used to identify natural disasters have made significant progress in recent years, which include machine learning, remote sensing techniques, and other information applications. Our literature review intends to provide a comprehensive overview of the various methods that are used for the classification of natural disasters and the various factors that influence their categorization.

Neural networks [2] offer a layered structure that allows for the extraction of complex features that cannot be obtained through traditional machine learning methods. When it comes to applications involving images, Convolutional Neural Networks (CNN) are often used due to their ability to perform image processing with ease. According to Tang et al. [3], CNN-based feature extraction is extremely effective and efficient, and CNNs are capable of identifying features even when the data is noisy.

Early works also considered using various models to classify the different types of natural disasters. Lucas Satria Aji et al. [4] suggested a technique for categorizing tweets as either disaster-related or non-catastrophic events relating to a crisis. The proposed methodology utilized a Convolutional Neural Network (CNN), having a Bidirectional Encoder Representation from Transformers (BERT) as a BERT Embedding. This approach was compared to another BERT embedding method called Word2Vec. Post-training and testing of the CNN model with BERT embeddings, their evaluation outcomes showed more accurate results with a precision score of 96.3%, a recall score of 96%, and an F1 score of 96.1%.

In [5], Tushar Agrawal et al. considered the model's training dataset to include images that are sorted into four categories: earthquakes, cyclones, wildfires, and floods. To detect disasters, the model receives either a live video stream or a pre-recorded video as input. It then examines each frame, calculating the probability of each class for that specific frame. The model selects the class with the highest probability as the label for that frame.

Data augmentation in disaster classification generates additional variations of the existing dataset, thereby providing the model with a more diverse set of images to learn from, resulting in improved accuracy and performance. Mummaneni Sobhana et al. [6] suggested a system that yielded compelling outcomes, achieving a 92% accuracy rate. The 9,600 images employed in training the model were found to be superior to the initial unaltered dataset, which produced an 88% accuracy rate. The model's remarkable enhancement was attributed to the additional orientations generated through the process of augmentation.

# 3 Proposed System and Materials

## 3.1 Dataset

To develop a machine-learning model for classifying natural disasters, a dataset comprising various images depicting floods, earthquakes, volcanoes, and wildfires is required. The dataset should consist of images with diverse color ranges and provide class labels for each disaster, along with a binary parameter for true or false. To fulfill our requirements for a dataset, abound in images, we will be using the 'Incidents1M Dataset' [7].

Incidents1M is an extensive dataset for discovering disasters, containing 977,088 images, 43 incidents, and 49 location categories. It is a transformation of the Incidents dataset that doubles the data size and contains more disaster tags.

## 3.2 Preparation

The initial dataset received, i.e., the 'Incidents1M Dataset' contained numerous types of incidents and disasters, however, the primary target of this paper is to detect natural disasters, i.e., floods, earthquakes, wildfires, and volcanoes. Initially, the dataset contained 43 separate incident labels, which were then segregated into the four required class labels for the model through a Python script that fetched all the URLs for the necessary classes.

Furthermore, images were scrapped using the in-built Python module 'requests' to get all the images downloaded in separate folders. Due to some of the URLs being expired, invalid, or blocked by the firewall, the final count of the images for each incident had a slight decrease in comparison to the original one.

## 3.3 Cleaning

The pre-cleaned dataset consisted of corrupted images, unnecessary images, negative class images, and various duplicate images for each disaster. The noisy image categories as mentioned were cleaned playing a crucial role in defining the final dataset used by the models.

Because the true number of images per class was required for training the model and having all true incident images in a particular class label, negative class images must be removed. Following that, unnecessary images that were irrelevant to the disasters must be manually removed, even though they were marked as the specific class labels in the dataset, which may affect the trained model's accuracy. To have accurate class labels for each disaster, more than 1,000 images per dataset were manually deleted.

Duplicate images can skew the results and waste computational resources. Hence, the removal of duplicate images is a crucial step in cleaning the dataset. An application named '4DDiG' [8] was used for the removal of duplicate images which were having at least 75% similarity between them. Furthermore, the use of a Python script with a Python package 'imagededup' [9] using its Perceptual Hashing (PHash) [10] to remove more duplicate images from the dataset.

Through the first two steps, most of the dataset was cleaned, however, there were some corrupted images left in the dataset that did not satisfy the image requirements such as '.jpeg', or '.png', and had to be removed. This is done with the help of a Python script using the module 'imghdr' [11], having the predefined extensions that are accepted in TensorFlow, such as '.jpeg','.png','.jpg', etc.

**Table 1.** Comparison of Image Counts Before and After Cleaning Of Different Natural Disaster Images

| **Disaster** | **Image Count** | |
|---|---|---|
| | ***Before*** | ***After*** |
| Flood | 11158 | 9924 |
| Wildfire | 8619 | 7534 |
| Earthquake | 10592 | 9512 |
| Volcano | 7720 | 6215 |
| **Total** | **38089** | **30185** |

After all the above cleaning steps were conducted, the final image count obtained totaled up to 30,185 images for all four disasters as illustrated in Table 1. With the 'flood' class label having the maximum number of images (9924) and Volcano having the least number of images (6215) post-cleaning of the dataset. This final dataset is now used for the final model training, with the use of CNNs, and ResNet.

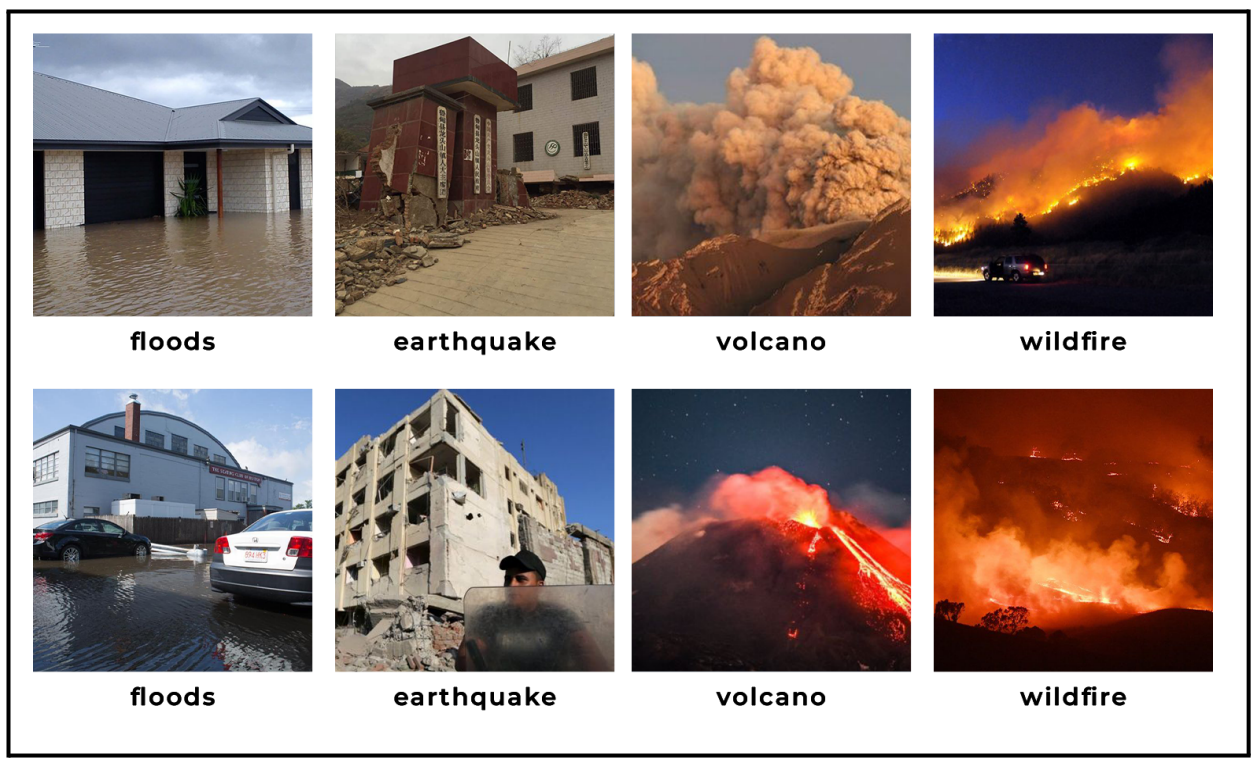

**Fig. 1.** Sample Images in the Dataset

# 4 Methodologies

## 4.1 Model 1 - CNN

Convolutional Neural Networks, also known as CNNs or ConvNets, are a subset of neural networks that process data with a familiar grid-like topology, such as a time series with a 1-dimensional structure or a typical 2-dimensional structure like Images. In recent times, Neural networks have gained a substantial amount of interest due to their structure and function being inspired by the human brain. It is primarily composed of a large number of interconnected nodes or neurons organized into layers.

Every node or neuron receives input from multiple other nodes or neurons from the previous layer in the network, processes the input received, and then outputs a signal to the nodes in the next layer of the network. Thus, ConvNets can be used to train on large datasets through a process called backpropagation [12].

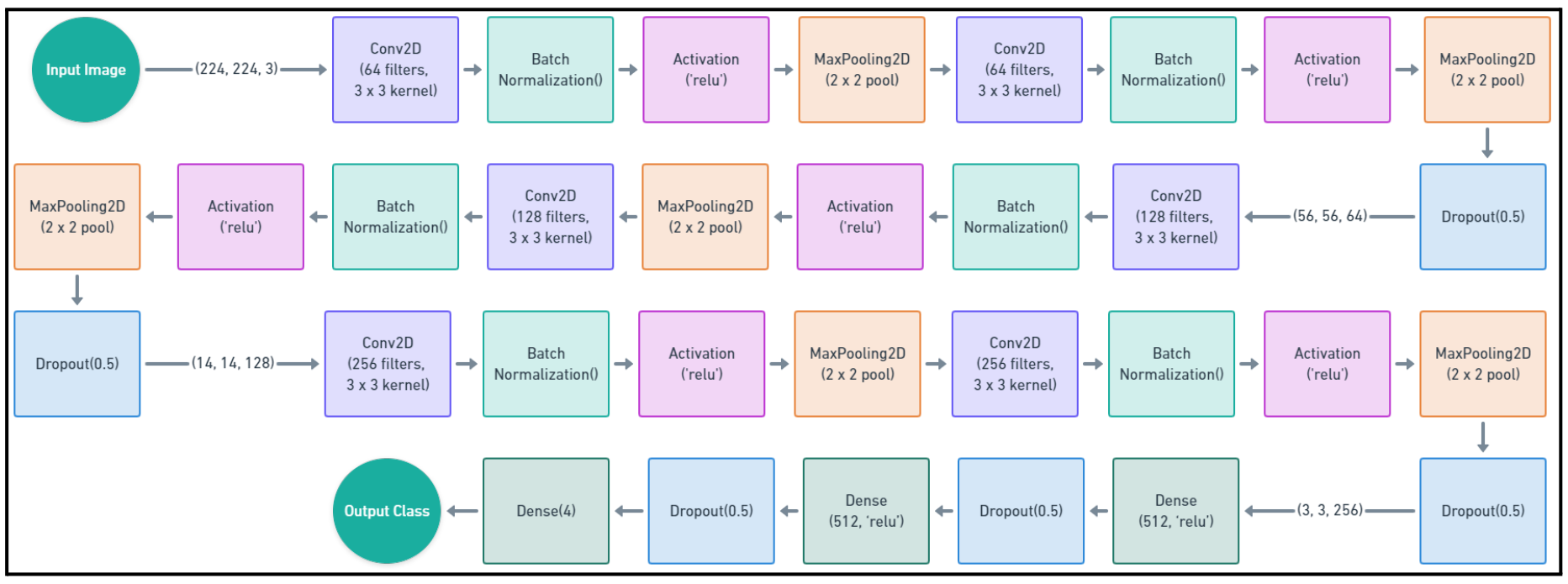

**Fig. 2.** Model 1 - CNN - Block Diagram

We leverage the use of CNNs for image classification. Since the dataset consists of images of various sizes and dimensions, as per Fig. 2 the model takes input images of size 224x224 pixels and rescales the pixel values to be between 0 and 1. Then we apply several convolution layers with activation functions and max pooling layers [13] to extract features from the images. Following this we applied the dropout layers for regularization and to prevent overfitting. To perform the final classification after flattening the feature maps, the network is then linked to fully connected layers. The output layer has 4 units, each unit corresponds to the 4 classes (earthquake, flood, volcano, wildfire) that the model can classify. Since our classification is a multi-class classification, we use the Sparse Categorical Cross Entropy Loss Function and finally compile the model with the Adam Optimizer [14]. The model's accuracy is also monitored during training and evaluation.

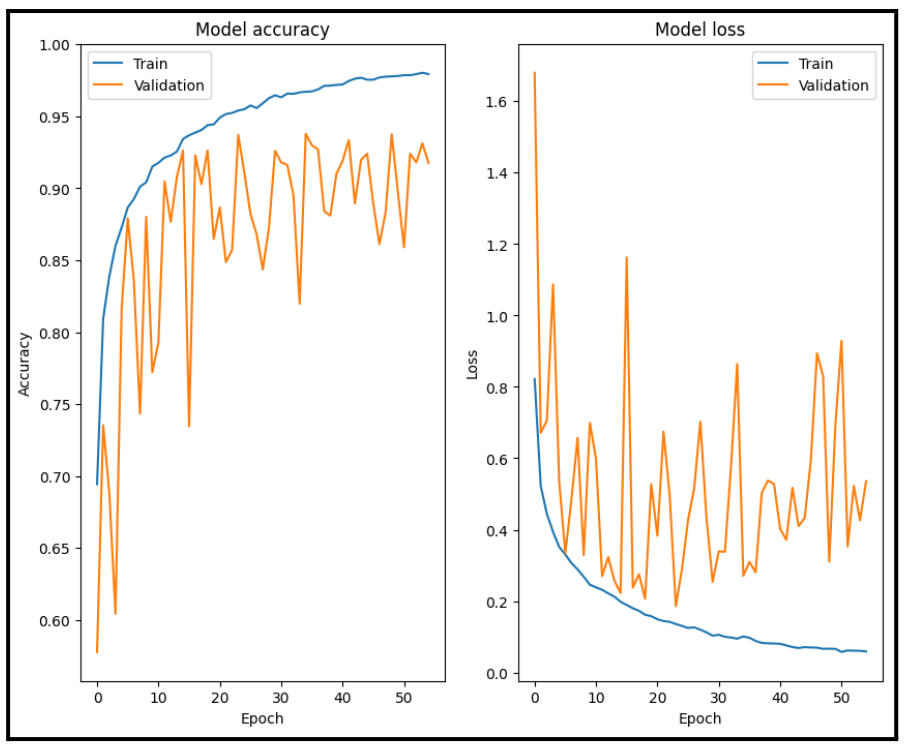

**Fig. 3.** Accuracy v/s Loss on train and test data (Model 1)

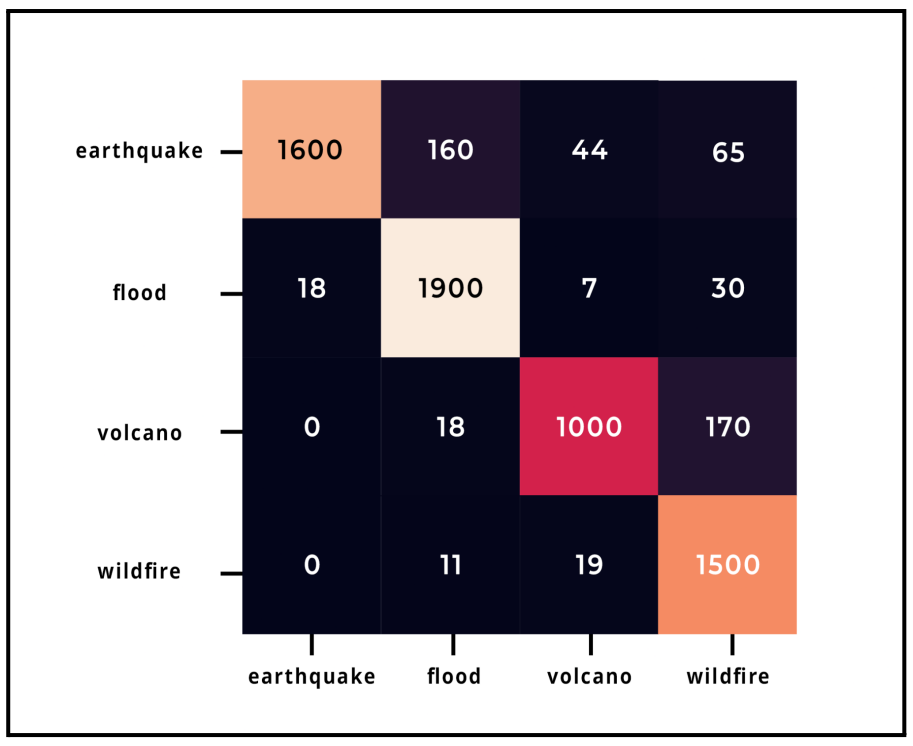

**Fig. 4.** Class Heatmap (Confusion Matrix) - (Model 1)

Given the complexity of the CNN layers and the hardware at our disposal, we initially trained with a batch size of 64 images being processed in a single backward and forward pass. Additionally, we set the epoch size to 90 to save the best CNN model that was trained. We noticed significant fluctuations in the Validation accuracy during the model's computation, which indicated an increase in epoch count and a decrease in batch size.

## 4.2 Model 2 - ResNet

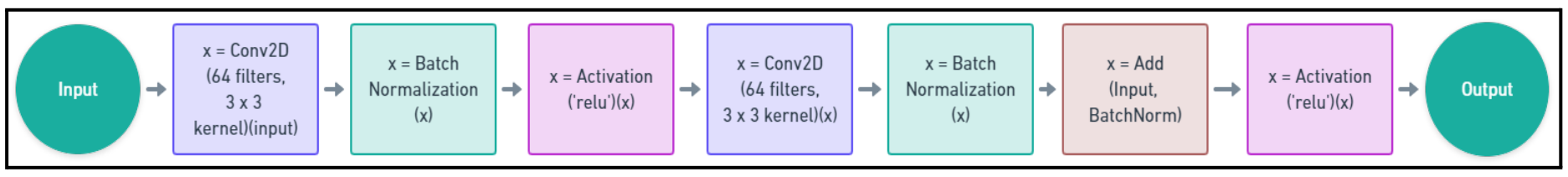

**Fig. 5.** Model 2 - ResNet - Single ResNet Block Diagram

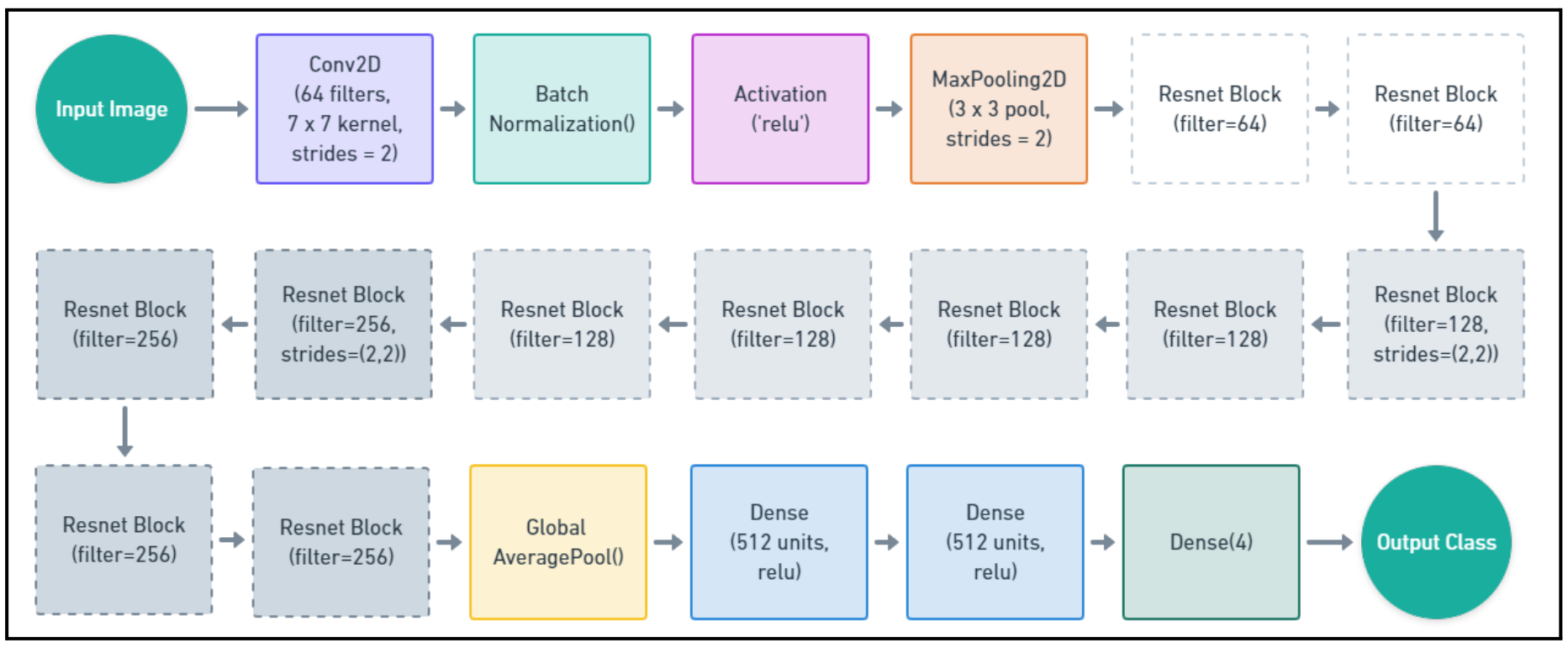

**Fig. 6.** Model 2 - ResNet - Block Diagram

ResNet is a Deep-learning architecture that is widely utilized for image recognition. It can effectively learn very deep neural networks using weight layers and residual functions. Here we propose to take advantage of the power of ResNet for the classification and Identification of Natural disasters. The proposed model is designed to classify images of disasters into various categories mentioned earlier. Our model incorporates the ResNet backbone followed by a set of fully connected layers for classification.

The foundation of our ResNet model is made up of several residual blocks, each with multiple convolutional layers, followed by activation functions, and batch normalization layers as shown in Fig. 5. Residual connections at each node allow direct data transfer from one layer to another, which helps us solve the problem of vanishing gradients and allows our model to efficiently learn the deep architecture.

As depicted in Fig. 6, the architecture involves the use of various ResNet blocks varying in the filter size from 64 to 256. The model takes input images of size 224x224 each with 3 channels and is processed through a Max Pooling Layer. When compared to using fully connected layers on top of convolutional features, the final convolutional layer's output proceeds through a Global Average Pooling layer, which helps to reduce overfitting. Next, two dense layers with the Rectified Linear Unit (ReLU) activation function are used [15], which outputs the input value if it is positive, otherwise 0, and densifies the network to 512 nodes. To perform the classification task, features learned from the ResNet trunk are mapped into four categories of disaster in the final fully connected layer, which is further dense into 4 nodes at the end of the model.

Based on previous model performance, we conducted experiments by varying the epoch count from 90 to 150 and adjusting the batch size from 64 to 16. As shown in Fig. 7, the highest accuracy was achieved when using an epoch count of 150 and a batch size of 16.

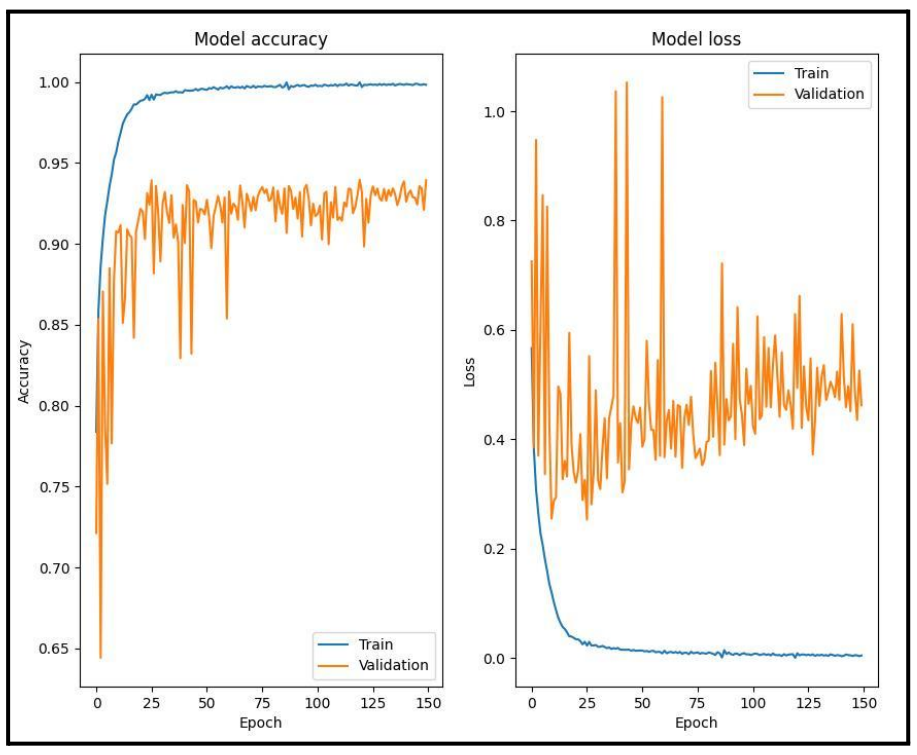

**Fig. 7.** Accuracy v/s Loss on train and test data (Model 2)

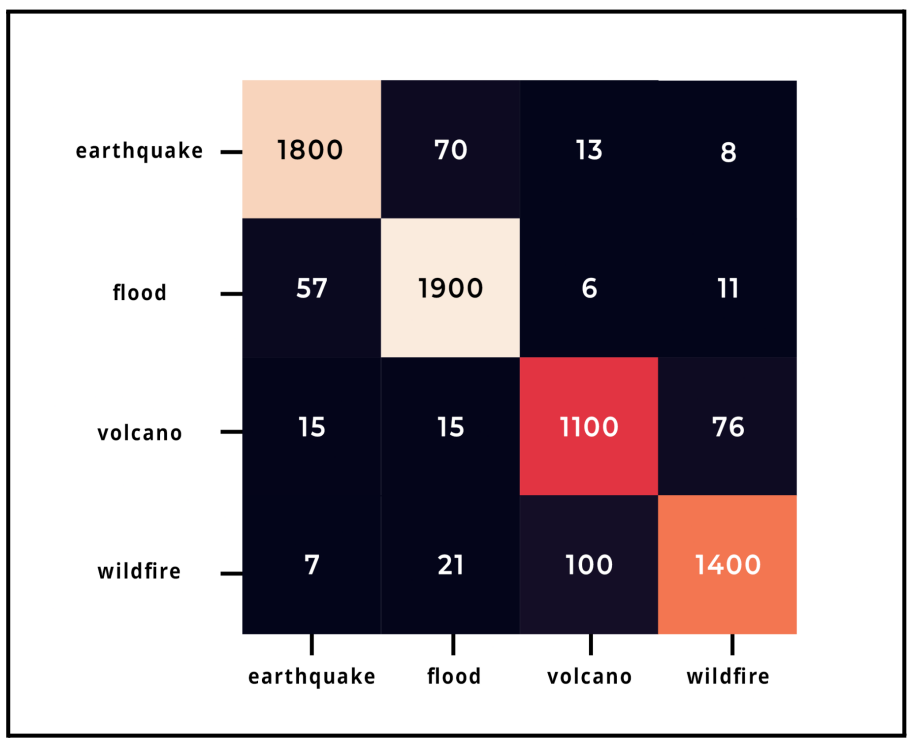

**Fig. 8.** Class Heatmap (Confusion Matrix) - (Model 2)

## 4.3 Model 3 - Stacking of CNNs

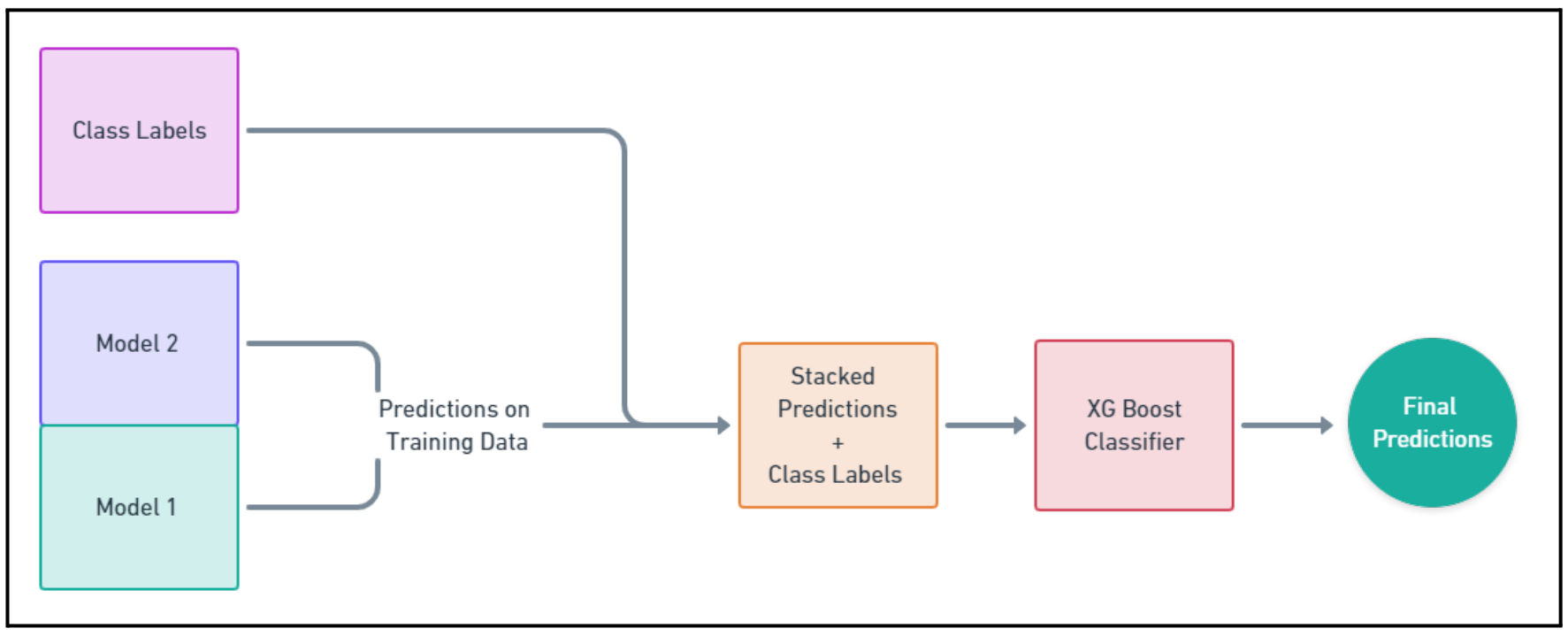

**Fig. 9.** Model 3 - Stacking of CNN - Block Diagram

Stacking of CNNs, also known as ensemble modeling or model stacking, is a technique in Machine Learning where we combine the predictions of multiple Convolutional neural Networks (CNNs) to improve the overall prediction performance. We leverage the strengths of individual models to make more accurate predictions collectively by blending or stacking previously trained models.

As per the block diagram shown in Fig. 9, we use the two earlier trained models on the Disaster Dataset which consists of labeled images namely, Model 1: CNN and Model 2: ResNet. Each model has generated predicted probabilities for the different classes for each image in the dataset. Now we stack the predicted probabilities of both models column-wise. Each row in the new dataset now contains the predicted probabilities from both CNN models and a column containing the true class labels of each image in the training dataset. The dataset, at this point, can be visualized as in Table 2.

**Table 2.** Stacked Dataset Of Predicted Probabilities from Model 1 And Model 2 (For All Classes)[a]

| Image | Model 1 - CNN | | | | Model 2 - ResNet | | | | True Classes |
|---|---|---|---|---|---|---|---|---|---|
| | *E*[b] | *F* | *V* | *W* | *E* | *F* | *V* | *W* | |
| 1 | 0.95 | 0.02 | 0.01 | 0.02 | 0.96 | 0 | 0.03 | 0.01 | E |
| 2 | 0 | 0.01 | 0 | 0.99 | 0.01 | 0 | 0 | 0.99 | W |

[a]This is a sample dataset created for understanding

[b]Symbols E, F, V, and W represent the four disasters

A tensorflow argmax() function is run on the dataset created in the previous step,, to find the class label with the highest predicted probabilities. Thus the stacked dataset can now be visualized as in Table 3.

**Table 3.** Stacked Dataset Of Predicted Probabilities from Model 1 And Model 2 (For Class With Highest Probability)[a]

| Image | Model 1 - CNN | Model 2 - ResNet | True Classes |
|---|---|---|---|
| 1 | Earthquake | Earthquake | E |
| 2 | Wildfire | Wildfire | W |

[a]This is a sample dataset created for understanding

The newly prepared stacked dataset now undergoes data modeling i.e., it is split into input features ‘X’ and the target labels ‘y’. Where ‘X’ contains the first two columns, which are the predicted class labels from Model 1 and Model 2. And ‘y’ contains the third column, which is the true class label from the validation dataset. Finally, after a train and test split we train the XGBoost [16] (Extreme Gradient Boosting) Classifier on the training dataset and further test on the test dataset to calculate the classification metrics and generate a classification report.

XGBoost Classifier is an ensemble learning method, meaning it combines the predictions from multiple models to make a final prediction. Here we train the XGBoost classifier on the predictions generated by Model 1 and Model 2. It uses a gradient-boosting strategy, which is an iterative process for building and improving weak learning models sequentially. Additionally, we used GridSearchCV from sci-kit learn, to perform a grid search over the specified hyperparameters like ‘max_depth', ‘learning_rate’ 'min_child_weight', and 'subsample' to find the best combination of hyperparameters that maximizes accuracy.

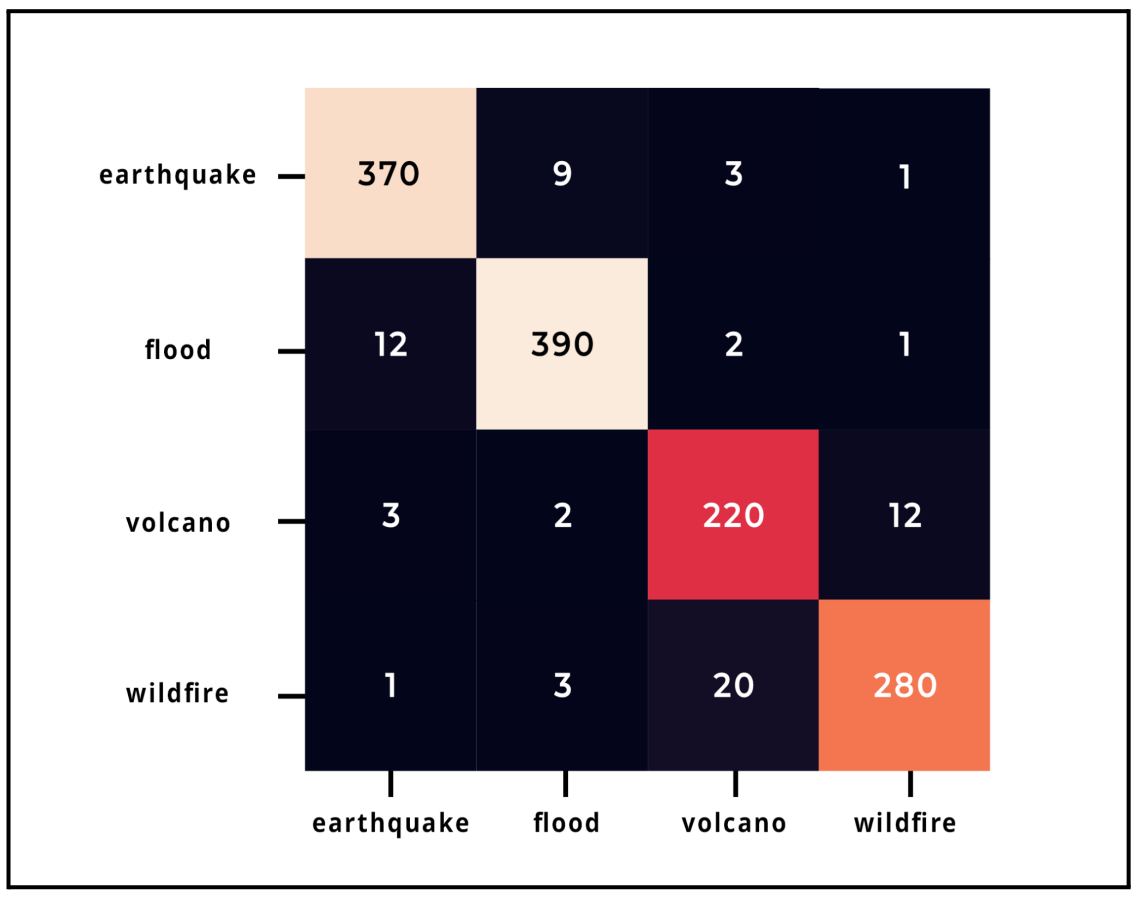

**Fig. 10.** Class Heatmap (Confusion Matrix) - (Model 3)

## 5 Experimental Results and Analysis

The experiment was carried out on a Windows computer with an RTX 3080 (12GB RAM GPU) and 32GB RAM. We used TensorFlow GPU [17] acceleration to train and test our CNN models by taking advantage of the GPU's (image-video memory) power. Setting up a conda environment and using the tensorflow-gpu [18] library to configure the physical devices were required steps in the setup process for TensorFlow GPU Acceleration. To maintain uniformity when performing analysis using the classification metrics, the image dataset obtained from the "Incidents 1M Dataset" from MIT University was cleaned and used in an 80/20 train-test split for the training and testing of all three models namely, Model 1 - Convolutional Neural Network, Model 2 - ResNet, and Model 3 - Stacking of CNNs.

The evaluation of the models was conducted by studying two frequently used parameters namely, Epochs and Batch Size, and a constant effort was made to optimize the two parameters to achieve the highest accuracy in classification.

**Epochs.** Each model was trained and optimized by using epochs. During the training process, an epoch is a single pass through the training dataset. To put it another way, one epoch means that the neural network has seen and processed every training example or image in the dataset once. Before training starts, the hyperparameter - number of epochs - needs to be defined. Our neural network workflow included a critical step in which determining the appropriate number of epochs required some experimentation and tuning. The following are some of the fundamental strategies used.

*Validation Curves.* The training and validation accuracy and loss curves are plotted as a function of epoch count. Initially, a decrease in the training and validation loss was noticed, but after a certain number of epochs, there were frequent hikes in the validation loss indicating overfitting. The optimal epoch level was estimated where the validation loss is at its minimum or where it starts to plateau.

*Checkpoints and EarlyStopping.* We used the EarlyStopping and checkpoint methodologies to save the state of the neural network model at various points during training. The models' architecture, parameters, and optimizer states are all included in each saved state. The primary reason we proposed this system was to enable us to stop training the model from a specific point in case of unexpected interruptions like system failures. During the training phase, EarlyStopping was used to monitor the validation loss and stop when it stopped improving. The model can then be restored to the checkpoint with the best validation performance.

**Batch Size.** Batch size was another hyperparameter that needed to be defined before the training process, similar to epochs. The term "batch size" simply refers to how many samples (images) are processed concurrently during training. The hardware, model, and dataset availability have a significant impact on batch size. More memory is required to store the activations and gradients during the forward and backward passes with a larger batch size. In general, we used the standard batch sizes, which are typically between 16 and 256, with values like 32, 64, and 128 being frequently used.

By using these techniques, we can ensure that we obtain the best possible neural network model without overfitting, thereby improving its generalization over images other than the dataset and real-world performance.

### 5.1 Evaluation of Model 1 - CNN

The base model for our disaster classification was a ConvNet with a learning rate of 0.001. The model was trained using 90 epochs and a batch size of 64 images processed in a single forward and backward pass. To stop training when the validation accuracy stopped improving, the model used EarlyStopping with a patience of 20 epochs. As the epoch number increased a noticeable plateau was formed in the validation accuracy curve, ranging from 85% to 93% and the EarlyStopping mechanisms stopped the training on the 55th epoch as shown in Fig. 3. A model checkpoint is established to save the best model based on the validation accuracy.

According to the classification report and from Fig. 4, the model was successfully trained with an accuracy of 92% and the highest precision in classifying Earthquakes.

### 5.2 Evaluation of Model 2 - Residual Neural Network (ResNet)

Our Residual Neural Network Model was built on the architecture of the Convolutional Neural Network - Model 1. It involved the use of various ResNet blocks, also known as Residual Blocks, which were stacked together to form the model's core, with different filter and stride values for down-sampling at specific points. With a learning rate of 0.0001, the model was compiled using an Adam Optimizer. Since disaster classification is a multi-class classification, we use the Sparse Categorical Cross Entropy Loss function to optimize the model.

We increased the epoch size from 90 to 150 epochs and reduced the batch size to 16 to reduce model run time and GPU load after thoroughly researching the previous model's performance. As per Fig. 7 even though the Validation Accuracy curve shows frequent fluctuations in the first 60 epochs, a plateau is formed from epoch 75 to 100 with an accuracy ranging from 90% to 94%. The best model based on validation accuracy is saved using Model Checkpoint.

According to the classification report and Fig. 8, the model was successfully trained with a 94% improvement in accuracy over the previous model. To further improve the ResNet model's ability to generalize to new images, we used Data Augmentation techniques. Data Augmentation [19] is a technique for increasing the size of a training dataset by generating new training images by transforming existing images.

We applied various TensorFlow methods such as 'RandomFlip,' 'RandomRotation,' 'RandomZoom,' and 'RandomContrast' to randomly flip images in horizontal and vertical directions, rotate and zoom images by a specific factor value, or adjust the contrast of an image by a random factor respectively. However, since this additional layer of data augmentation did not yield the expected results, it was skipped.

### 5.3 Evaluation of Model 3 - Stacking of CNNs

The third model, Stacking of CNNs, was created by stacking the previous two models, which served as base models, column by column. The next step in the stacking process involves the selection of a meta-model. A meta-model is trained and used to perform predictions, on the predictions obtained from the base models as depicted in the Block Diagram Fig. 9. We initially chose a Multiclass Logistic Regression as the meta-model, but we saw no improvement in classification accuracy. We then trained an XGBoost Classifier on the stacked dataset. The starting learning rate was set at 0.01. However, the XGBoost Classifier is optimized by the GridSearchCV [16] which is a Cross Validation technique that ensures to search for the best hyperparameters to get the best accuracy.

According to the classification report represented by Table 4., the model was successfully trained with a higher accuracy than the previous base models. Stacking produced predictions with 95% accuracy along with the highest precision in flood classification.

**Table 4.** Model Comparison Using Various Classification Metrics

| Model | Class | *Precision* | *Recall* | *$F_1$ Score* | *Support* | *Accuracy* |
|---|---|---|---|---|---|---|
| CNN - Model 1 | *earthquake* | 0.99 | 0.86 | 0.92 | 1907 | **0.92** |
| | *flood* | 0.91 | 0.97 | 0.94 | 2004 | |
| | *volcano* | 0.94 | 0.84 | 0.89 | 1219 | |
| | *wildfire* | 0.85 | 0.98 | 0.91 | 1507 | |
| ResNet - Model 2 | *earthquake* | 0.96 | 0.95 | 0.96 | 1907 | **0.94** |
| | *flood* | 0.95 | 0.96 | 0.96 | 2004 | |
| | *volcano* | 0.90 | 0.91 | 0.91 | 1219 | |
| | *wildfire* | 0.94 | 0.91 | 0.92 | 1507 | |
| Stacking of CNNs - Model 3 | *earthquake* | 0.96 | 0.97 | 0.96 | 387 | **0.95** |
| | *flood* | 0.97 | 0.96 | 0.96 | 403 | |
| | *volcano* | 0.90 | 0.93 | 0.91 | 237 | |
| | *wildfire* | 0.95 | 0.92 | 0.94 | 301 | |

Subsequent figure; Fig. 11 depicts a comparison of the three models based on the $F_1$ Score classification metric in each category of classification.

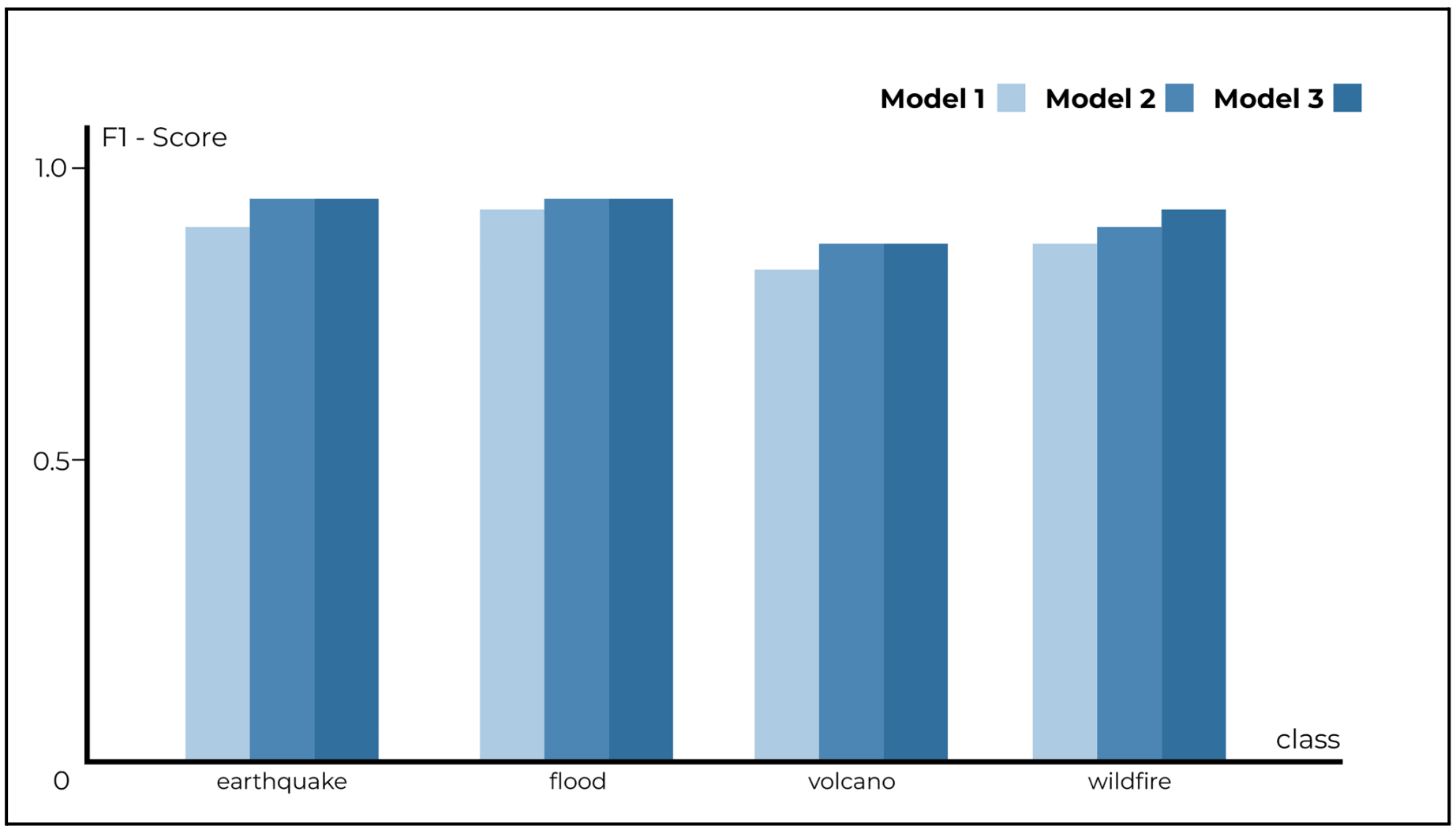

**Fig. 11.** $F_1$ Score of Classification (CNN v/s ResNet v/s Stacking)

## 6 Conclusion and Future Work

By performing this research work we have demonstrated a novel approach for classifying disaster images using Convolutional Neural Networks (CNNs). The results generated by our CNN-based models including Resnet and stacked CNN ensembles, have achieved an accuracy of over 95% in classifying key disasters namely earthquakes, floods, volcanoes and wildfires. As per our analysis, the stacked CNN ensemble model was built using the basic CNN and a Resnet Architecture as base models and finally Ensembling using XGBoost Classifier has emerged as the top performer. We can state that combining and leveraging the strengths of individual base models to create a new hypermodel helps in producing better results. Other than that optimal model tuning and selecting the best hyperparameters like learning rate, number of epochs and batch size were also important to maximize accuracy. Our research is well-established on the concepts of deep learning and CNNs for disaster image classification. The techniques we propose can be further utilized in the development of automated systems for disaster response, damage assessment as well and recovery management. Additionally a scope of disaster classification can be road segmentation. This involves providing survivors with a precise road or exit path after a catastrophe has occurred, thereby improving post-disaster recovery efforts. Despite the usage of large disaster imagery dataset, there are certain limitations. The dataset may not represent the full diversity of disasters in real life. Additional tuning might further improve the model performance. Future research should concentrate on increasing the size and diversity of the dataset, combining disaster image segmentation with classification, and deploying these models in real-world disaster management systems. Model robustness and generalization testing will also be essential. More robust disaster management systems can be developed to mitigate risks and improve preparedness globally by addressing these limitations and building on current CNN advances. This study lays the groundwork for progress towards that goal.